\documentclass{article}
\usepackage{spconf,amsmath}
\usepackage{subfigure}
\usepackage{url}
\usepackage{graphicx}
\graphicspath{{Figures/}}
\newcommand{\tabincell}[2]{
\begin{tabular}{@{}#1@{}}#2\end{tabular}}

\title{Foreground Detection in Camouflaged Scenes}
\name{Shuai Li\textsuperscript{*}, Dinei Florencio\textsuperscript{\dag}, Yaqin Zhao\textsuperscript{\ddag}, Chris Cook\textsuperscript{*}, Wanqing Li\textsuperscript{*}\thanks{Shuai Li is partially supported by the Global Challenge Project, Assistive Systems for the Ageing, of University of Wollongong.}}
\address{University of Wollongong\textsuperscript{*}, Microsoft Research\textsuperscript{\dag}, Nanjing Forestry University\textsuperscript{\ddag}}

\begin{document}

\maketitle

\begin{abstract}
Foreground detection has been widely studied for decades due to its importance in many practical applications. Most of the existing methods assume foreground and background show visually distinct characteristics and thus the foreground can be detected once a good background model is obtained. However, there are many situations where this is not the case. Of particular interest in video surveillance is the camouflage case.  For example, an active attacker camouflages by intentionally wearing clothes that are visually similar to the background. In such cases, even given a decent background model, it is not trivial to detect foreground objects. This paper proposes a texture guided weighted voting (TGWV) method which can efficiently detect foreground objects in camouflaged scenes. The proposed method employs the stationary wavelet transform to decompose the image into frequency bands. We show that the small and hardly noticeable differences between foreground and background in the image domain can be effectively captured in certain wavelet frequency bands. To make the final foreground decision, a weighted voting scheme is developed based on intensity and texture of all the wavelet bands with weights carefully designed. Experimental results demonstrate that the proposed method achieves superior performance compared to the current state-of-the-art results.
\end{abstract}

%
\section{Introduction}
Detection of moving foreground objects is a crucial step in many vision-based systems. Most foreground detection algorithms are based on background subtraction. More specifically, they compute a background model and make foreground decisions by comparing the current image against that background model. A classic example (and still one of the most popular) of a background subtraction method is the Gaussian Mixture Model (GMM) \cite{kaewtrakulpong2002improved} and the improved adaptive Gaussian Mixture Model \cite{zivkovic2004improved,zivkovic2006efficient} known as ``MOG'' and ``MOG2'', respectively, which uses a few Gaussian distributions to model the intensity of each pixel. A pixel-based adaptive method which, instead of using the distribution of pixels, uses a history of recently observed pixels for background modeling was proposed in \cite{hofmann2012background}, noted as ``SubSENSE''. While all these methods provide useful classification for many situations, they all use only the intensity of the current pixel, and will likely fail when the pixels of the background and the foreground object share similar color.

Another class of algorithms \cite{heikkila2006texture, st2014flexible, han2012density, jian2008background, varadarajan2009background, benedek2008bayesian}, instead of looking simply at the pixel intensity, tries to detect foreground objects based on features that intend to capture the texture around each pixel, i.e., the relationship between the current pixel and a small predefined neighborhood. These features include  ``Haar'', ``Gradient'' and ``Local Binary Pattern (LBP) \cite{ojala2002multiresolution}''. However, these kinds of features can only exploit a certain type of texture in a small predefined region and cannot adaptively change the scale. When a pixel is located in a region which has relative poor texture at the chosen resolution, the method will likely fail. There have been a few methods \cite{huang2003wavelet,huang2004double,kushwaha2015framework} in the literature using wavelet transform for background subtraction. However, the existing methods only use the wavelet transform as an additional feature without considering the characteristics of wavelet transform, and thus do not take full advantage of the wavelet transform.

\begin{figure}[tb]
\centering
\subfigure{
\includegraphics[width = 0.30\hsize]{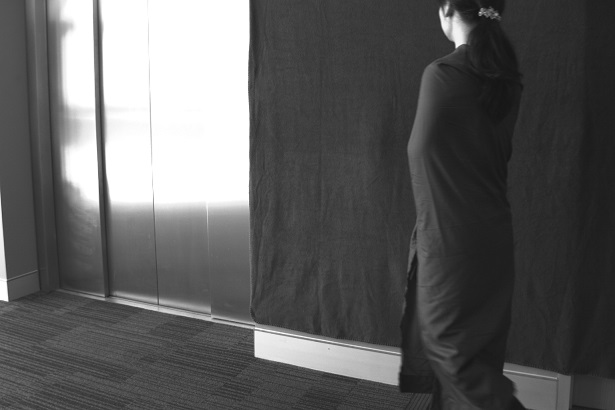}}
\subfigure{
 \includegraphics[width = 0.30\hsize]{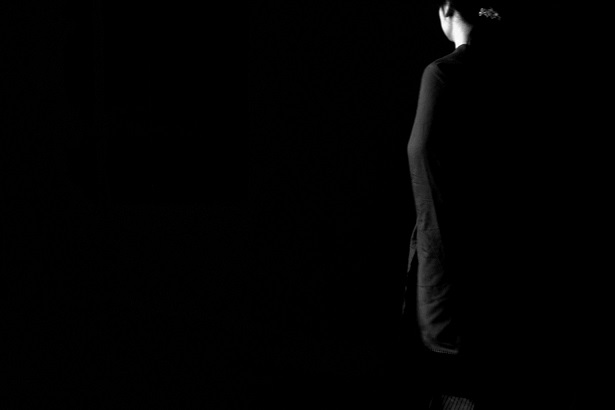}}
\subfigure{
 \includegraphics[width = 0.30\hsize]{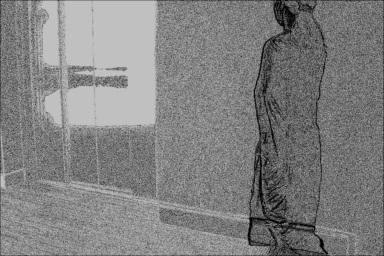}}
\caption{Example of a camouflaged scene and its differences in terms of intensity and LBP in the image domain between the current image and the background image.}
\label{CamExample}
\end{figure}

A comprehensive review of recent background subtraction and foreground detection methods can be found in \cite{sobral2014comprehensive}. Existing methods were developed to deal with general foreground objects that show distinct intensity or texture changes. However, there are situations where foreground objects may share similar intensity and texture as the background, such as camouflaged scenes, especially those with poor texture. For example, a person may wear clothes that share similar color to that of the background wall as shown in Fig. \ref{CamExample}. These cases pose great challenges to foreground detection but are extremely important for surveillance applications, but they have not yet received much attention in the literature.

This paper focuses on the foreground detection for camouflaged scenes, addressing the above-mentioned problem where foreground objects are visually similar to the background and lack distinct intensity or texture differences. The contributions of this paper are summarized as follows.
\begin{itemize}
\item We investigate the differences between foreground objects and visually similar background in the image and wavelet domain, and show that the small differences in the image domain may be able to be detected in certain wavelet frequency bands.
\vspace{-0.2cm}
\item We propose a weighed voting method to combine all the decisions made on the differences in terms of intensity and texture of the bands, respectively.
\vspace{-0.2cm}
\item Three weights are carefully designed, namely noise-induced weight, texture-guided weight and translation weight, to deal with noise, different characteristics and scales of different frequency bands, respectively.
\vspace{-0.2cm}
\item A new dataset for camouflaged scenes captured in real life is collected with labeled groundtruth. Experiments on both the existing dataset (artificially generated) and the new dataset demonstrate significantly better performance using the proposed method over the state-of-the-art methods.
\end{itemize}

\vspace{-0.2cm}
\begin{figure}[t]
\begin{center}
\begin{tabular}{c@{}c@{}c@{}c@{}}
\includegraphics[width = 0.25\hsize]{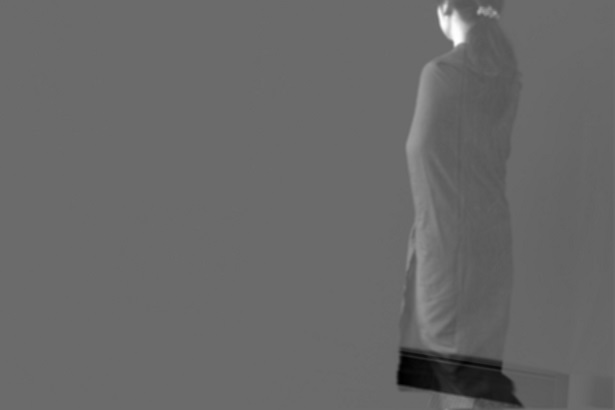} & \includegraphics[width = 0.25\hsize]{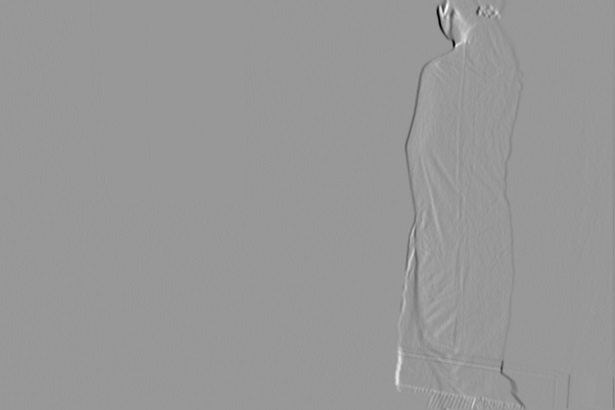}& \includegraphics[width = 0.25\hsize]{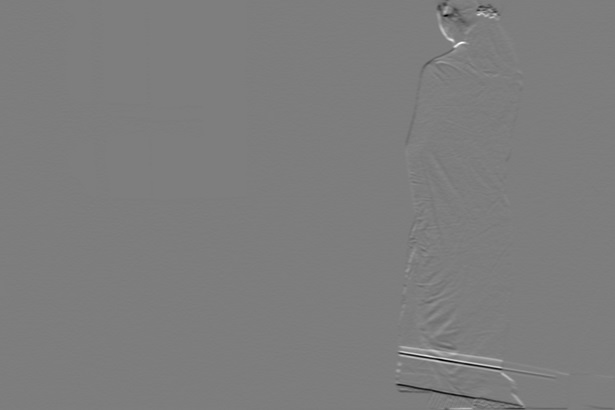}& \includegraphics[width = 0.25\hsize]{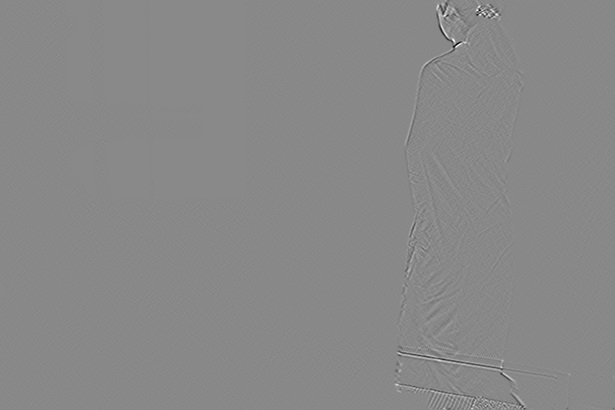}\\
\includegraphics[width = 0.25\hsize]{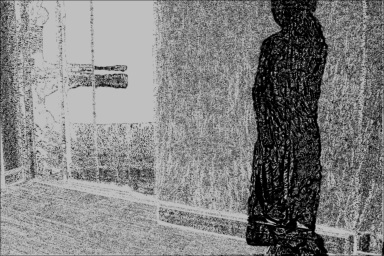} & \includegraphics[width = 0.25\hsize]{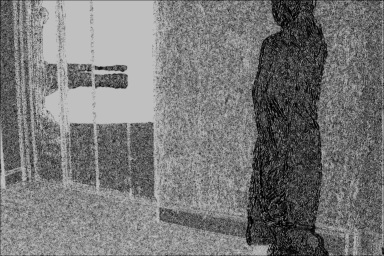}& \includegraphics[width = 0.25\hsize]{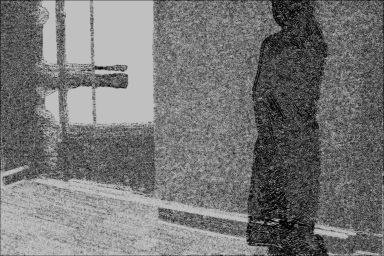}& \includegraphics[width = 0.25\hsize]{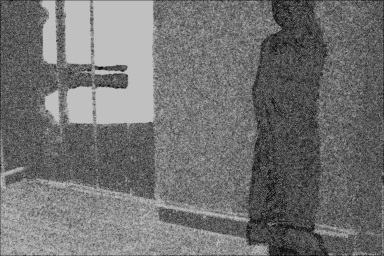}\\
LL Band & LH Band &HL Band&HH Band\\
\end{tabular}
\end{center}
\caption{Differences in terms of intensity (wavelet coefficients) and texture (LBP) between the current image and the background image in each wavelet band of the 3-rd level.}
\label{WaveletDiff}
\end{figure}

\section{Foreground Detection in The Wavelet Domain}
\vspace{-2mm}
In foreground detection, much effort has been spent on obtaining an appropriate background model but detecting the foreground from the background model is still largely based on the differences between the current image and the background image. Although this may work for scenes where foreground object and background show distinct changes in intensity and texture, it cannot work appropriately for camouflaged scenes where foreground objects are visually similar to background. The first image in Fig. \ref{CamExample} gives an example of a person wearing clothes that share similar color as the wall. The corresponding differences in the image domain in terms of intensity and texture (LBP) between the foreground and background (obtained by MOG2) are shown in Fig. \ref{CamExample} where the values in the image are properly scaled for display. It can be seen that the differences between the foreground person and background in the camouflaged scene are really small and consequently directly detecting the foreground in the image domain may fail.

\subsection{Intensity and Texture Analysis in Wavelet Domain}
Wavelet transform decomposes images into different frequency bands, allowing for multiresolution analysis. Specially, each stage of a wavelet transform decomposes an image into four frequency bands, typically denoted as LL, LH, HL, and HH. The LL band contains the low frequency approximation of the image, while the other three bands contain the details (high frequency) of the image in the horizontal, vertical and diagonal directions, respectively. This paper employs a stationary wavelet transform \cite{nason1995stationary,pesquet1996time}, i.e., a non-decimated redundant wavelet transform. The Haar wavelet basis is used, as it provides  a simple and effective basis. Considering that we only make use of the wavelet bands as features instead of reconstructing the original image, all wavelet bands including the low frequency wavelet bands at all levels are used in the proposed method.

Different levels of wavelet decomposition show different frequency bands and each frequency band extracts different characteristics of an image. Small differences in the image domain may be highlighted in one or a few wavelet bands when large differences in certain frequencies exist between foreground objects and the background. Fig. \ref{WaveletDiff} shows the intensity (wavelet coefficient) differences and the texture (LBP) differences in the wavelet domain, respectively, in each band of level 3 (other levels are very similar and thus not shown here). Compared to the difference in the image domain shown in Fig. \ref{CamExample}, it can be seen that the differences become apparent in the wavelet domain, which makes the detection of the foreground from the visually similar background possible.

\vspace{-1mm}
\subsection{Texture Guided Weighted Voting (TGWV)}
\begin{figure}[t]
\centering
\includegraphics[width = 1\hsize]{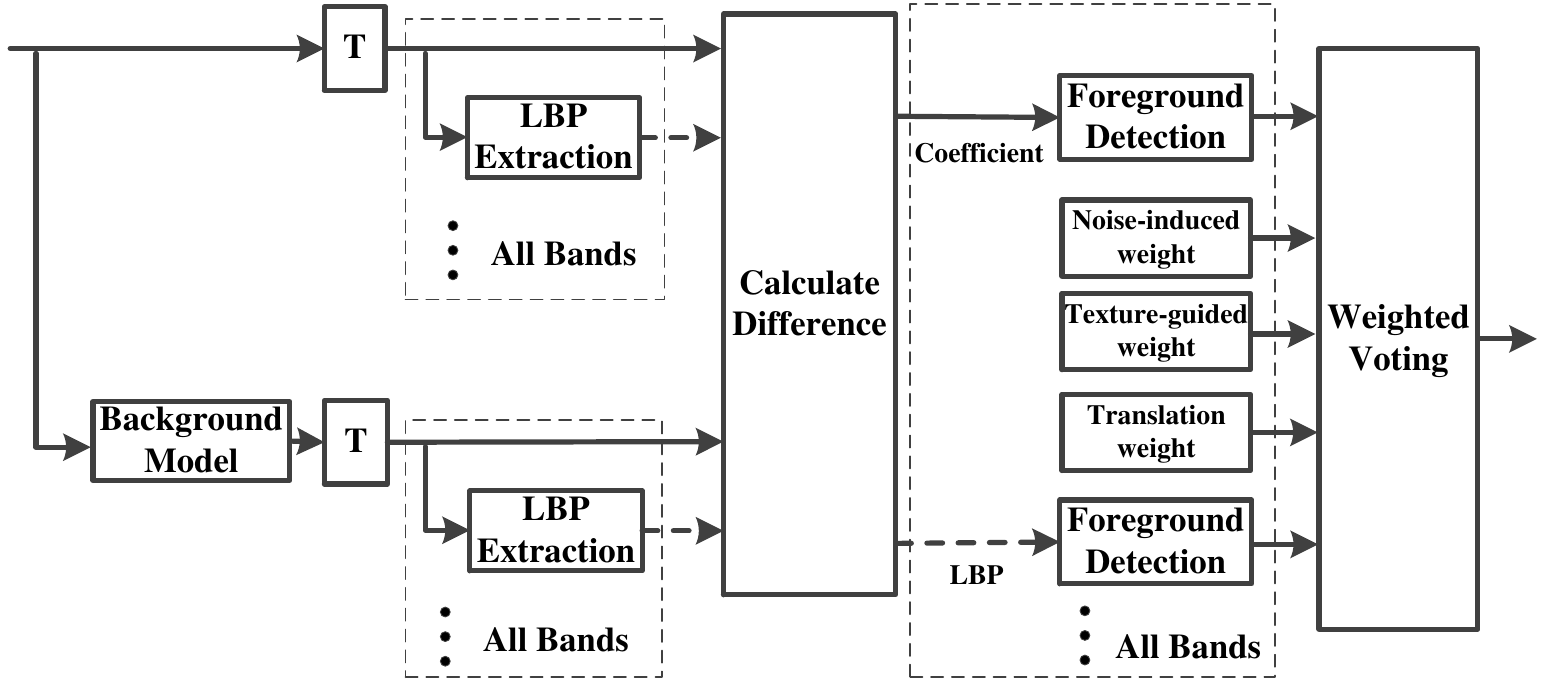}
\caption{Framework of the foreground detection method developed in this paper.}
\label{diagram}
\end{figure}

The framework of the TGWV based foreground decision method developed in this paper is shown in Fig. \ref{diagram}. First a background model is constructed and the background image for the current time is formed. As this paper mainly focuses on foreground detection for camouflaged scenes, for simplicity existing background modeling processes such as the GMM in MOG2 \cite{zivkovic2006efficient} and the SuBSENSE in \cite{st2014flexible} are used. We have evaluated in the Experimental Results Section that our proposed foreground detection method can be efficiently combined with any background modeling method while achieving superior performance. Then the stationary wavelet transform is employed to decompose the background image and the current image into wavelet bands of $M$ levels ($M$ is empirically set as 7 in the current experiments). The uniform LBP histogram is then extracted for each wavelet band to represent their textures, referred as LBP hereafter. The differences between the current image and the background image in the wavelet domain are obtained as the absolute differences of the coefficients and the LBP differences measured by the histogram intersection kernel \cite{swain1991color}. Then foreground decisions are made separately for the coefficients and LBP in each wavelet band as in GMM by formulating a Gaussian distribution on the differences. Finally the decision for each pixel is obtained by a weighted voting scheme which will be explained in the following.

\subsubsection{Noise-induced Weight}
It is known that noise exists in images, even in uncompressed ones, coming from the acquisition process. Usually white Gaussian noise, which has zero mean and shows the same energy at different frequency bands, is assumed. The decisions made on each wavelet band may be affected by the noise and thus the confidence level of each decision could be compromised. However, energies of different wavelet bands are quite different and thus the effects of the noise on the decisions are different. Since that the decision on each wavelet band is based on the value of the coefficient, the confidence $\omega _{ni}$ of the decision by considering the effect of the noise can be determined as 
\begin{small}
\begin{equation}
\omega _{ni} =\frac{\sigma _{si} - \sigma _n}{\sigma _{si}}
\label{eq:noise_weight}
\end{equation}
\end{small}
where $\sigma _{si}$ and $\sigma _n$ are the standard deviation of the $i$-th wavelet band and noise, respectively.

\subsubsection{Texture-guided Weight}
Different frequency bands show different responses to different textures. For example, high frequency wavelet bands only respond to the changes across pixels, which makes them not discriminative for flat regions. Therefore, to adaptively adjust the confidence of the low frequency and high frequency wavelet bands in each level on the final decision for different regions with different textures, a texture guided weight $w_t$ is used by taking into consideration the texture (measured by LBP) of the region that the current pixel located in. 
\begin{small}
\begin{align}
\omega_{tLi} &= 1 - f\left.((n_i(BG)+n_i(C))\middle/ K\right.) \nonumber\\
\omega_{tWi} &= \begin{cases}
1 - f\left.((n_i(BG)+n_i(C))\middle/ K\right.), \hspace{1mm}if ~ i \in \{LH, HL, HH\};\\
1 + \sum_{k \in \{LH, HL, HH\}}2 \cdot f\left.((n_{k}(BG) + n_{k}(C)\middle / K)\right.) \\
\hspace{2.5mm}+ f\left.((n_i(BG)+n_i(C))\middle/ K\right.), ~ if ~ i = LL
\end{cases}
\label{eq:Weights_bands}
\end{align}
\end{small}
where $\omega_{tLi}$ and $\omega_{tWi}$ represent the weights for the decisions made on LBP and coefficients, respectively. $n_i(BG)$ and $n_i(C)$ represent the number of the last two patterns in the uniform LBP (the pattern with no changes and the pattern with more than two changes indicating how flat the region is) in the background image and the current image, respectively, and $K$ is the total number of all the patterns. $f(x)$ is a non-decreasing function that maps $x$ to a value between $0$ and $1$. It can be seen that when the current region is very flat, the foreground detection of the wavelet bands in this level will be determined mostly based on the low frequency wavelet band information.

\subsubsection{Translation Weight}
As each wavelet band can be considered as a representation of the image in a certain frequency, final foreground decisions of pixels can be obtained by effectively combining all the decisions together. Considering that each wavelet coefficient is obtained based on a few pixels in a block, a translation process is needed to transfer the decisions in the wavelet domain to the image domain. It is clear that if all the pixels related to a coefficient belong to the same object, the decision made on this coefficient can well represent the decision on these pixels in this frequency. On the contrary, if pixels related to a coefficient belong to different objects, the decision made on this coefficient may not be correct for all these pixels. In order to characterize this relationship, the correlation among pixels is first modeled using a first-order autoregressive process as 
\begin{small}
\begin{equation}
\rho (i+1) = \alpha \cdot \rho(i) + \epsilon(i+1)
\label{eq:}
\end{equation}
\end{small}
where $\alpha$ , $0<\alpha<1$, is the autoregressive coefficient and $\epsilon$ is a white noise process with zero mean. When the distance between two pixels increases, their correlation reduces. Therefore, as the wavelet decomposition level increases and the number of pixels related to one coefficient increases, the correlation between the central and the related pixels gets smaller as their distance apart get greater. Consequently, the confidence of the decision made on the coefficient being correct to all its related pixels gets smaller. A translation weight $\omega _{ci}$ is used for each wavelet coefficient to transfer the decision to its related pixels, which is determined as the average correlation of all the pixels related to the coefficient. 
\begin{small}
\begin{equation}
\omega _{ci}=\frac{1}{N_p}\sum_{i=1}^{N_p} {\rho(i)}
\label{eq:translationweight}
\end{equation}
\end{small}
where $N_p$ is the total number of the pixels related to one coefficient.

\subsubsection{Weighted Voting based Foreground Detection}
By combining the noise induced weight in (\ref{eq:noise_weight}), the texture guided weight in (\ref{eq:Weights_bands}) and the translation weight in (\ref{eq:translationweight}), we can obtain the final decision for each pixel using a weighted voting strategy as
\begin{small}
\begin{align}\label{eq:WeightedVote}
V &=\sum_{i=1}^{N}{\omega_{ni} \cdot \omega_{ci} \cdot (\omega_{tWi} \cdot V_{Wi} + \omega_{tLi} \cdot V_{Li})}\nonumber\\
F &=\begin{cases}
1, \hspace{4mm} if \hspace{2mm} V > T_V; \hspace{2mm}\\
0, \hspace{4mm} otherwise
\end{cases}
\end{align}
\end{small}
where $V_{Wi}$ and $V_{Li}$ represent the foreground decisions made based on the wavelet coefficient and LBP in the $i$-th wavelet band, respectively, and $N$ represents the total number of wavelet bands. $F$ represents the final foreground decision for a pixel, and $T_V$ is the minimum number of votes needed to determine the current pixel as foreground. 

\begin{figure*}[t]
\centering
\subfigure{
 \includegraphics[width = 0.32\hsize]{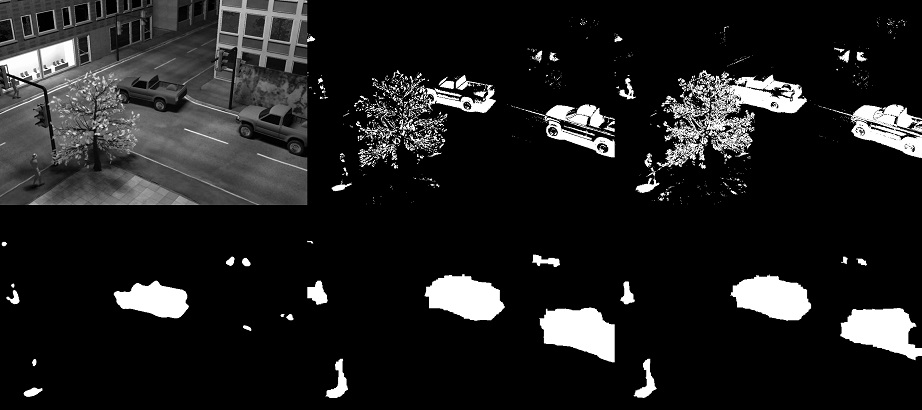}}
\subfigure{
\includegraphics[width = 0.32\hsize]{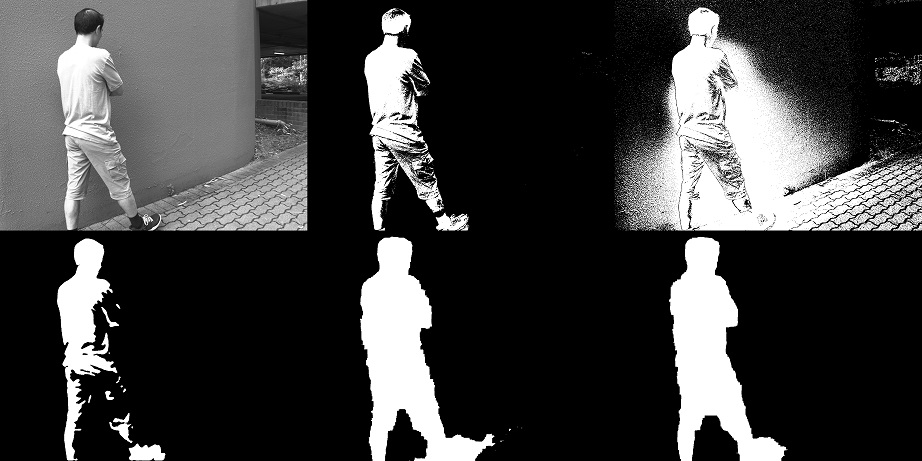}}
\subfigure{
 \includegraphics[width = 0.32\hsize]{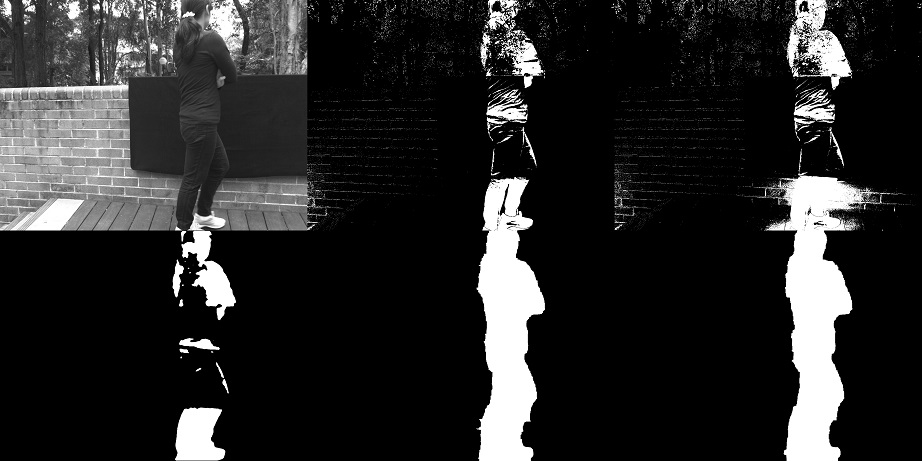}}
\subfigure{
 \includegraphics[width = 0.32\hsize]{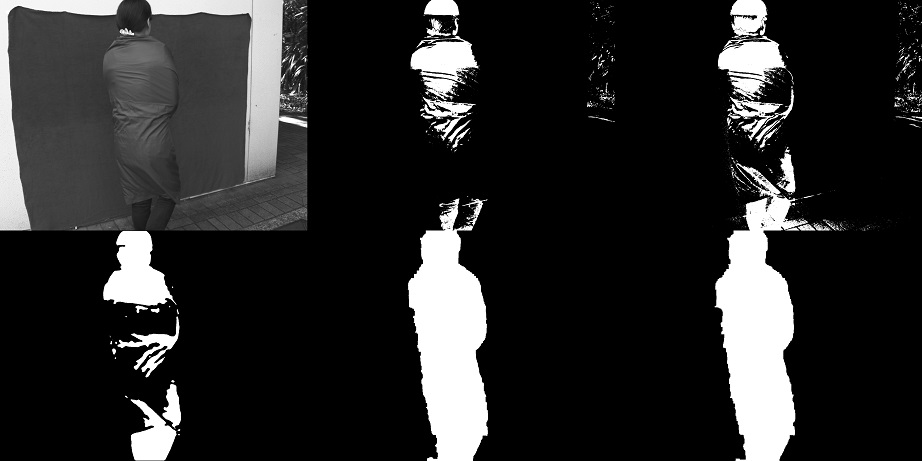}}
\subfigure{
 \includegraphics[width = 0.32\hsize]{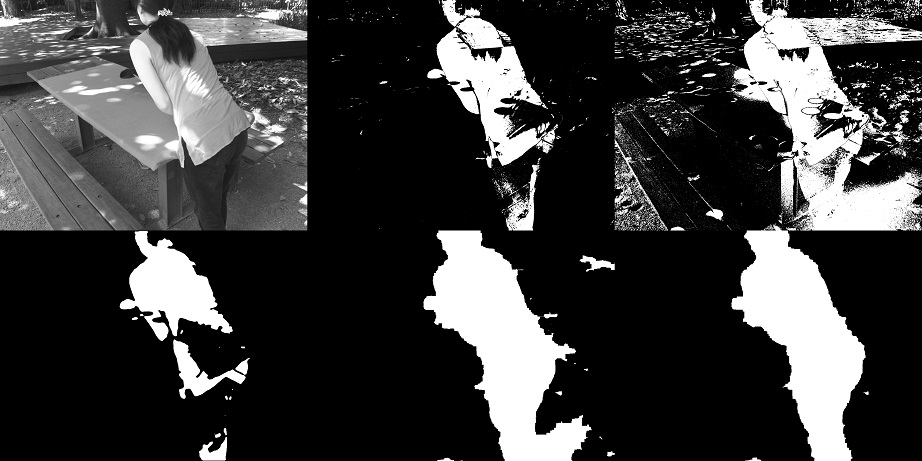}}
\subfigure{
 \includegraphics[width = 0.32\hsize]{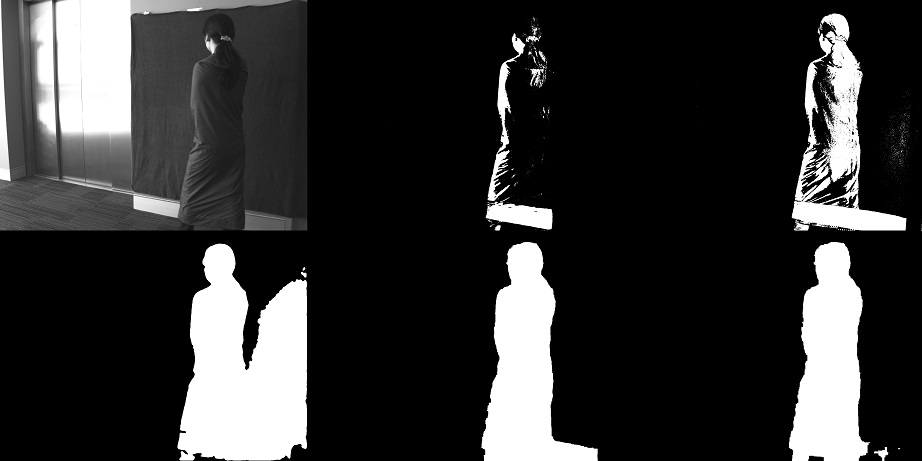}}
\caption{Detection result examples. For each figure, the images in the first row from left to right are the current image, the results obtained by ``MOG2 \cite{zivkovic2006efficient} without shadow'', ``MOG2 \cite{zivkovic2006efficient} with shadow'', respectively. The images in the second row from left to right are the results obtained by ``SubSENSE\cite{st2014flexible}'', ``Proposed + MOG2'' and ``Proposed + SuBSENSE'', respectively.}
\label{SubjectiveExperiment}
\end{figure*}

\vspace{-3mm}
\section{Experiments}
\vspace{-5mm}
\subsection{Datasets}
\vspace{-3mm}
As reported in \cite{brutzer2011evaluation}, a camouflaged video has been used to test the performance of the existing algorithms for detecting camouflaged objects, which is adopted in this paper for evaluation. However, this dataset was artificially generated by computer graphics and it can hardly simulate the texture on the real foreground object. Therefore, we captured another dataset containing 5 grayscale video sequences of resolution 1536*1152 in real scenes. The foreground person wears clothes in a similar color as that of the background. Examples of the scenes with the foreground person in the collected dataset together with the camouflaged video \cite{brutzer2011evaluation} are shown in Fig. \ref{SubjectiveExperiment} (first image of each figure).

\vspace{-5mm}
\subsection{Results}
\vspace{-3mm}
The proposed foreground detection method is compared with the popular MOG2 \cite{zivkovic2006efficient} and one most recent work SubSENSE \cite{st2014flexible} which has reported the state-of-the-art performance on background subtraction. The performance is measured using Recall, Precision, F-Measure and PSNR obtained by the BMC Wizard software \cite{BMCWizard}. Due to the page limitation, only the average experimental results of all the videos including the camouflaged video in \cite{brutzer2011evaluation} are shown as in Table \ref{aveResult}. As expressed in the previous Section, the proposed method focuses on detecting the foreground from the visually similar background and the background modeling process is implemented using the GMM model in MOG2 \cite{zivkovic2006efficient} or SubSENSE \cite{st2014flexible} which are denoted by ``Proposed + MOG2'' and ``Proposed + SuBSENSE'', respectively. From the results, it can be seen that the proposed method can achieve significant improvement. Especially in terms of Recall, the proposed method can perform much better and can detect most of the foreground objects. It can be also seen that regardless of the background modeling methods used, the proposed method consistently achieves superior performance. Some example detection results are also shown in Fig. \ref{SubjectiveExperiment}. It can be clearly seen that the results obtained by the proposed method are better than the existing methods in detecting the foreground objects that are visually similar to the background.

\begin{small}
\begin{table}[tbp]
\centering
\caption{Average Performance of All Tested Videos}
\label{aveResult}
\begin{tabular}{|p{5.7em}|c|c|c|c|}
\hline
Method & Recall & Precision & F-Measure & PSNR\\
\hline
\tabincell{l}{MOG2\cite{zivkovic2006efficient}\\with shadow} & 0.859	& 0.731 &	0.788 &	30.162\\
\hline
\tabincell{l}{MOG2\cite{zivkovic2006efficient}\\without shadow} & 0.786	& 0.879	& 0.826	& 36.792\\
\hline
SuBSENSE\cite{st2014flexible} & 0.847 &	0.911	& 0.873	& 38.994\\
\hline
\tabincell{l}{Proposed\\+ MOG2} & 0.984 &	0.876 &	0.926	& 42.904\\
\hline
\tabincell{l}{Proposed\\+ SuBSENSE} & 0.979 &	0.899	& 0.937	& 45.645\\
\hline
\end{tabular}
\end{table}
\end{small}


The computational complexity of the proposed method is higher than MOG2 \cite{zivkovic2006efficient} and SubSENSE \cite{st2014flexible} due to the multiple level wavelet decomposition. However, the algorithm is highly parallel and can be easily implemented on GPU \cite{franco2009parallel}.

\vspace{-3mm}
\section{Conclusion}
\vspace{-3mm}
This paper presents a texture guided weighted voting scheme in the wavelet domain to addresses the foreground detection problem in camouflaged scenes. It is first shown that small and hardly noticeable differences between foreground objects and visually similar background can be well captured by a few wavelet bands. Then foreground decisions are made for each wavelet band based on wavelet coefficients and local binary patterns obtained on the coefficients, respectively. To effectively combine all the decisions for the final detection on the pixels, the property of the wavelet transform is considered and a texture guided weight is adaptively determined for each decision. Experimental results have verified the efficacy of the proposed foreground detection method.

\bibliographystyle{IEEEbib}
\bibliography{Reference}

\end{document}